\documentclass[letterpaper, 10 pt, conference]{ieeeconf}  %

\IEEEoverridecommandlockouts                              %

\usepackage{graphics} %
\usepackage{epsfig} %
\usepackage{times} %
\usepackage{amsmath} %
\usepackage{amssymb}  %
\usepackage{multirow}
\usepackage{tikz}
\usetikzlibrary{spy}
\usepackage{pgfplots}
\usepackage{pgfplotstable}
\usepgfplotslibrary{groupplots}
\usepackage{booktabs}
\usepackage{subcaption}
\usepackage{lipsum}
\usepackage{makecell}
\usepackage{tabularx}
\usepackage{xcolor}
\usepackage[pagebackref=true,breaklinks=true,bookmarks=false,colorlinks=true]{hyperref} %

\definecolor{1COLOR}{cmyk}{90, 90, 0, 0}
\definecolor{2COLOR}{cmyk}{0, 90, 0, 0}
\definecolor{4COLOR}{cmyk}{0, 90, 90, 0}
\definecolor{5COLOR}{cmyk}{0, 0, 0, 90}
\definecolor{6COLOR}{cmyk}{90, 0, 0, 0}

\definecolor{barrier}{RGB}{255,120,50}
\definecolor{bicycle}{RGB}{255,192,203}
\definecolor{bus}{RGB}{255,255,0}
\definecolor{car}{RGB}{0,150,245}
\definecolor{construction_vehicle}{RGB}{0,255,255}
\definecolor{motorcycle}{RGB}{200,180,0}
\definecolor{pedestrian}{RGB}{255,0,0}
\definecolor{traffic_cone}{RGB}{255,240,150}
\definecolor{trailer}{RGB}{135,60,0}
\definecolor{truck}{RGB}{160,32,240}
\definecolor{driveable_surface}{RGB}{255,0,255}
\definecolor{other_flat}{RGB}{139,137,137}
\definecolor{sidewalk}{RGB}{75,0,75}
\definecolor{terrain}{RGB}{150,240,80}
\definecolor{manmade}{RGB}{230,230,250}
\definecolor{vegetation}{RGB}{0,175,0}

\definecolor{LightCyan}{rgb}{0.87,0.92,0.96}
\definecolor{m_green_border}{HTML}{82B366}
\definecolor{m_darkgreen}{HTML}{d5e9d5}
\definecolor{m_green}{HTML}{d5e9d5}
\definecolor{m_orange}{HTML}{fff2cd}
\definecolor{m_red}{HTML}{fccecd}
\definecolor{m_violet}{HTML}{e2d6e8}
\definecolor{m_blue}{HTML}{d8e9fc}
\definecolor{m_red}{RGB}{255,209,209}
\definecolor{m_red_border}{RGB}{215,23,20}
\definecolor{m_orange_border}{RGB}{150,114,164}
\definecolor{m_blue_border}{RGB}{107,141,190}
\definecolor{m_yellow}{RGB}{255,242,205}
\definecolor{m_yellow_border}{RGB}{213,182,82}
\definecolor{m_gray}{RGB}{245,245,245}
\definecolor{m_gray_border}{RGB}{102,102,102}

\newcommand{\colorsquare}[1]{\tikz{\path[draw=#1_border,fill=#1, thick, rounded corners=0.6pt] (0,0) rectangle (6pt,6pt);}}

\newcommand{\PAR}[1]{\vskip4pt \noindent {\bf #1~}}

\renewcommand*\backref[1]{\ifx#1\relax \else #1 \fi}

\newcolumntype{K}[1]{>{\centering\arraybackslash}p{#1}}

\pgfplotsset{
    1_style/.style={color=1COLOR, dashdotted, mark=triangle*, thin, mark size=2pt},
    2_style/.style={color=2COLOR, solid, mark=halfdiamond*, thin, mark size=2pt},
    4_style/.style={color=4COLOR, densely dashdotted, mark=diamond*, thin, mark size=2pt},
    5_style/.style={color=5COLOR, dashed, mark=star, thin, mark size=2pt},
    6_style/.style={color=6COLOR, solid, mark=halfcircle*, thin, mark size=2pt},
}

\newif\ifmynotes
\mynotestrue
\definecolor{notetext}{rgb}{0.7,0,0}

\title{\LARGE \bf
OCCUQ: Exploring Efficient Uncertainty Quantification for \\ 3D Occupancy Prediction
}

\author{Severin Heidrich$^{1*}$, Till Beemelmanns$^{1*}$, Alexey Nekrasov$^{2*}$, Bastian Leibe$^{2}$, Lutz Eckstein$^{1}$%
\thanks{This work has received funding from the European Union’s Horizon Europe Research and Innovation Programme under Grant Agreement No. 101076754 - AIthena project. The project was partially funded by the BMBF project “WestAI” (grant no. 01IS22094D)}%
\thanks{$^{1}$The author is with the Institute for Automotive Engineering, RWTH Aachen, 52074 Aachen, Germany {\tt\small \{firstname.lastname\}@ika.rwth-aachen.de}}%
\thanks{$^{2}$The author is with the Computer Vision Institute (i13), 52074 Aachen, Germany {\tt\small \{lastname\}@vision.rwth-aachen.de}}%
\thanks{$*$-denotes equal contribution}
}

\begin{document}

\maketitle
\thispagestyle{empty}
\pagestyle{empty}

\begin{abstract}
Autonomous driving has the potential to significantly enhance productivity and provide numerous societal benefits.
Ensuring robustness in these safety-critical systems is essential, particularly when vehicles must navigate adverse weather conditions and sensor corruptions that may not have been encountered during training.
Current methods often overlook uncertainties arising from adversarial conditions or distributional shifts, limiting their real-world applicability.
We propose an efficient adaptation of an uncertainty estimation technique for 3D occupancy prediction.
Our method dynamically calibrates model confidence using epistemic uncertainty estimates.
Our evaluation under various camera corruption scenarios, such as fog or missing cameras, demonstrates that our approach effectively quantifies epistemic uncertainty by assigning higher uncertainty values to unseen data.
We introduce region-specific corruptions to simulate defects affecting only a single camera and validate our findings through both scene-level and region-level assessments.
Our results show superior performance in Out-of-Distribution (OoD) detection and confidence calibration compared to common baselines such as Deep Ensembles and MC-Dropout.
Our approach consistently demonstrates reliable uncertainty measures, indicating its potential for enhancing the robustness of autonomous driving systems in real-world scenarios. 
Code and dataset are available at \url{https://github.com/ika-rwth-aachen/OCCUQ}.
\end{abstract}

\section{Introduction}
Autonomous driving represents a pivotal advancement with the potential to significantly increase productivity and provide numerous benefits to society.
A critical aspect of this field is understanding and ensuring the robustness of autonomous systems, especially in such a safety-critical application.
In real-world scenarios, vehicles must navigate and respond to various challenges posed different weather conditions and potential sensor corruptions.
Adverse weather conditions such as rain, fog, or even sensor malfunctions such as a faulty camera can occur during deployment and may not have been encountered during training.

Current methods for autonomous driving use a multi-camera setup to construct 3D occupancy maps.
These maps consist of voxels representing the space occupancy, different semantic classes, and serve as input for the trajectory planning and collision avoidance stack~\cite{trajectoryplanning}.
Ensuring the reliability and accuracy of these maps under diverse conditions is essential for the safe and effective operation of autonomous vehicles.

In our work, we focus on developing a 3D occupancy prediction research direction.
Current approaches focus on generating datasets for 3D occupancy prediction~\cite{wang2023openoccupancy, tian2024occ3d,wei2023surroundocc} and improvements in the model architecture~\cite{wei2023surroundocc,huang2023tpvformer,li2023fb,li2023voxformer}, but often overlook uncertainties that may arise from adversarial conditions or distributional shifts.
This oversight significantly hampers their deployment in real-world scenarios.

\begin{figure}
    \centering
    \includegraphics[width=0.8\linewidth]{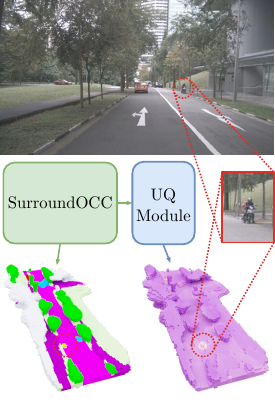}
    \caption{
    Our method introduces a lightweight Uncertainty Quantification Module to the main model, that provides uncertainties for 3D occupancy fields.
    Motorcyclists in the distance has higher uncertainty scores indicating it as a rare class in the training set.
    }
    \label{fig:teaser}
\end{figure}

Understanding and quantifying uncertainty is crucial because it helps identify situations where perception models might fail, thus enabling the implementation of fail-safe mechanisms, active learning strategies~\cite{activelearning}, or other relevant downstream tasks~\cite{wang2023uqapplication}.
Uncertainty estimation for autonomous driving has been a growing field in recent years~\cite{feng2020review}, with multiple benchmarks~\cite{acdc, beemelmanns2024multicorrupt, kong2023robo3d} proposed to evaluate various methods.
Currently, multiple works~\cite{cortinhal2020salsanext,cao2024pasco} apply Deep Ensembles~\cite{lakshminarayanan2017deepensemble}, MC-Dropout~\cite{gal2016dropout} or derivative methods~\cite{mimo, masksembles, valdenegro2019deepsubensemble} to obtain uncertainty estimates.
In practice, these methods require long inference times or increase the memory footprint significantly, making them difficult to apply in practice.
However, no investigation was proposed for efficient uncertainty estimation in 3D occupancy prediction.

In recent years, multiple works have explored the potential to replace costly uncertainty quantification methods with more efficient ones~\cite{masksembles, valdenegro2019deepsubensemble}.
A recent line of work~\cite{van2020uncertainty,mukhoti2023ddu} suggests the possibility of a very efficient uncertainty estimation without multiple forward passes or increased memory requirements.
Direct uncertainty estimation methods provide accurate uncertainty estimations and introduce little overhead in practice.
However, these methods have not yet been investigated for 3D occupancy prediction.

In our work, we focus on the adaptation of an efficient uncertainty estimation method for 3D occupancy prediction.
By incorporating an uncertainty module~\cite{mukhoti2023ddu} in the dense 3D occupancy detection head and separately training a Gaussian Mixture Model (GMM) at the feature level, we aim to disentangle aleatoric and epistemic uncertainty during inference.
To evaluate uncertainty values, a common approach in the community is to assess Out-of-Distribution (OoD) detection performance.
We evaluate our method for the case of different camera corruptions~\cite{beemelmanns2024multicorrupt}, such as fog or a missing camera.
However, there are no datasets in autonomous driving, that can provide input corruptions at voxel level.
To evaluate the ability of our method to localize corruptions, we introduce region-specific corruptions by simulating defects affecting only a single camera.
To confirm our findings, we inspect uncertainty values for individual objects that are expected to have higher epistemic uncertainty.
In addition, we use the epistemic uncertainty estimate to dynamically calibrate the confidence of the model.
Our results suggest that techniques for efficient uncertainty estimation provide reliable uncertainty measures at scene and region level.
Our approach demonstrates superior performance across various downstream tasks related to uncertainty estimation, including scene-level, region-level OoD detection and confidence calibration. 

Our main contributions are as follows: \textbf{(1)} We propose a simple and efficient adaptation of an uncertainty estimation technique to the case of 3D occupancy prediction; \textbf{(2)} we compare our approach for OoD detection and confidence calibration with common uncertainty baselines; \textbf{(3)} We investigate design choices for the integration of the uncertainty quantification module.

\begin{figure*}[ht]
    \centering
    \includegraphics[width=0.95\linewidth]{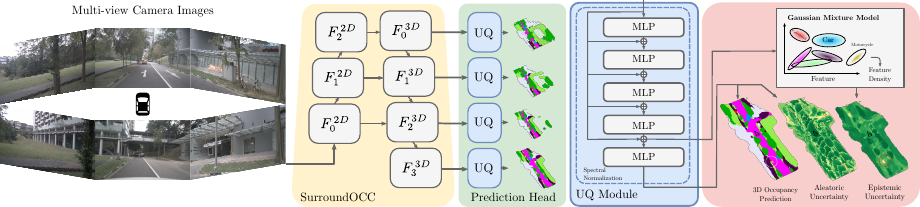}
    \caption{
    \textbf{Overview of the proposed method.}
    From multi-view camera images our method provides 3D occupancy predictions with reliable epistemic and aleatoric uncertainties at voxel level.
    We build on top of the SurroundOCC~\cite{wei2023surroundocc} model, and introduce an uncertainty quantification (UQ) module.
    }
    \label{fig:method_overview}
\end{figure*}

\section{Related Work}

\subsection{Uncertainty Estimation}
The are two main sources of uncertainty, aleatoric (data) and epistemic (model) uncertainty~\cite{gal2016uncertainty}.
Epistemic uncertainty is a measure of the lack of knowledge of the model, and is typically used to solve a downstream task, such as Out-of-Distribution Detection~\cite{lakshminarayanan2017simple}, while aleatoric uncertainty is used to provide the evaluation of the amount of noise in the data.
These two uncertainties cannot be disentangled in many cases, and the overall predictive uncertainty will be high whenever either epistemic or aleatoric uncertainty is high~\cite{mukhoti2023ddu}.
Common baselines for uncertainty estimation include Deep Ensembles (DE)~\cite{lakshminarayanan2017deepensemble} and Monte-Carlo Droupout~\cite{gal2016dropout}.

Some recent approaches focus on an improvement over MC Dropout and Deep Ensembles, making uncertainty estimation during inference faster and requiring less memory~\cite{durasov2021masksembles,valdenegro2019deepsubensemble,laurent2023packedensembles,mimo,huang2017snapshot}.
Recently, a parallel direction focusing on uncertainty estimation in a single forward pass~\cite{Amersfoort2020SimpleAS,liu2020simple,vanamersfoort2022featurea} gained attention.
By estimating the densities of the representation space, it is possible to infer uncertainties~\cite{mukhoti2023ddu}.
In our work, we explore the application of an efficient uncertainty estimation replacing only a fraction of the model.

Uncertainty estimation in autonomous driving spans from planning and motion forecast~\cite{xu2014motion,malinin2022shifts} to uncertainty-aware semantic segmentation~\cite{cortinhal2020salsanext,acdc}. %
Recently, in the field of 3D semantic scene completion from LiDAR, PaSCo~\cite{cao2024pasco} proposed to use MIMO technique~\cite{mimo} for uncertainty estimation.
However, no works explore uncertainty estimation techniques for occupancy prediction from images.

\subsection{3D Occupancy Prediction}
The task of 3D occupancy prediction has recently gained attention for its ability to capture fine-grained semantic geometric information for improved 3D scene understanding, compared to object-level detections.
Recent approaches can be classified into three categories: BEV-based, TPV-based, and voxel-based methods \cite{zhang2024vision}.

In BEV-based methods, 3D information is projected into a 2D bird’s-eye view where the height is collapsed, focusing on the ground plane.
These methods are computationally efficient.
FB-OCC \cite{li2023fb}, for example, uses Lift-Splat-Shoot (LSS) \cite{philion2020lift} to lift 2D image features into 3D space and compress them into a BEV feature map, followed by backward projection for refinement.
OccFormer \cite{zhang2023occformer} uses a dual-path transformer to process local and global features, while OpenOccupancy \cite{wang2023openoccupancy} introduces a benchmark with a two-step process for coarse and refined occupancy prediction.

TPV-based methods extend BEV by adding two additional perpendicular planes for richer spatial representation.
TPVFormer \cite{huang2023tri} generalizes MonoScene \cite{cao2022monoscene} for multi-camera setups, using cross-view attention in a transformer-based architecture to enable interactions among the three planes.
However, it remains limited by its sparse LiDAR supervision, resulting in less detailed predictions.

MonoScene \cite{cao2022monoscene} was the first voxel-based method, projecting 2D image features into 3D space, but struggled with single-view limitations.
VoxFormer \cite{li2023voxformer} addresses sparsity by using a two-stage process with depth-based voxel queries and transformer-based feature enhancement.
CTF-Occ \cite{tian2024occ3d} uses a transformer-based model to aggregate 2D image features into 3D space via voxel queries and cross-attention.
SurroundOcc \cite{wei2023surroundocc}, chosen for this study, efficiently uses 2D-3D attention to map low-resolution features to a voxel grid, refining granularity via 3D deconvolution for dense, surround-view occupancy prediction.

\section{Method}
To address the challenging problem of uncertainty estimation in autonomous driving, we explore an efficient uncertainty estimation method in application to 3D occupancy prediction. 
As a base model, we leverage SurroundOCC~\cite{wei2023surroundocc} for 3D occupancy prediction from multi-camera surround view images.
Inspired by Deep Deterministic Uncertainty~\cite{mukhoti2023ddu}, we introduce an uncertainty module that allows uncertainty estimation at voxel level.

\PAR{Overview.}
Input to the model are $6$ multi-view camera images, one from the front and the back, and two cameras for each side.
These images are processed by a multi-resolution feature backbone~(Fig.~\ref{fig:method_overview}~\colorsquare{m_orange}) to produce a dense feature volume. 
The feature volume is processed by a prediction head~(\colorsquare{m_green}) to predict the occupancy of a voxel and a semantic class.
In our work, we replace the prediction head with our lightweight uncertainty module~(\colorsquare{m_blue}).
In addition to semantic and occupancy prediction, the uncertainty module encapsulates a Gaussian Mixture Model that is used to predict the model's epistemic uncertainty.
The output of the model is a set of occupancy and semantic predictions, as well as the aleatoric and epistemic uncertainties at voxel level~(\colorsquare{m_red}).
We describe the parts of the model in the following sections.

\PAR{Feature Backbone.}
Our method uses SurroundOCC~\cite{wei2023surroundocc} as its core component to construct a 3D feature volume $F_n^{\,3D} \in \mathbb{R}^{H \times W \times Z \times C}$ from 2D images $I = \{I_{0},I_{1},..,I_{6}\}$.
SurroundOCC uses a ResNet backbone to process multi-view images to extract multi-scale features for each camera.
The transformation from 2D feature volume to 3D is performed using 3D queries from every position in the feature volume.
These 3D voxel queries are projected to the 2D views using intrinsic and extrinsic camera parameters, updated with the 2D information using deformable cross-attention mechanism~\cite{dai2017deformable}. 
To improve the information content of neighboring voxels, a 3D convolution is applied to the constructed volume, followed by a deconvolution operation as the final stage to obtain a dense feature volume.
The resulting representation is enriched by incorporating features from lower resolutions, thus following a 2D-3D UNet network architecture. 
Prediction heads at each resolution are supervised with voxel occupancy and semantic classes.

\PAR{Prediction Heads and Uncertainty Module.}
For uncertainty estimation, we introduce an uncertainty-aware modification to the prediction heads~(Fig.~\ref{fig:method_overview}~\colorsquare{m_green}).
Inspired by the DDU method~\cite{mukhoti2023ddu}, we use feature density as an uncertainty measure of the model.
To ensure that the feature density correlates well to the uncertainty, a well-regularized feature space is necessary.
Our model can be expressed as a logit function $V(I) = g \circ f_\theta(I)$ with the backbone ${f_\theta: I \rightarrow \mathcal{H}}$, that maps input images $I$ to a feature representation $f_\theta(I) \in \mathcal{H}$, and an output function $g$ that maps $f_\theta(I)$ to the label space.
We aim to ensure that the distance in the feature space $D_{\text{feature}}(f_\theta(I) - f_\theta(I'))$ corresponds to the distance in the input space $D_{\text{input}}(I - I')$ between the seen and unseen data.
To make our model distance aware, two conditions are crucial:
\begin{enumerate} 
  \item \textbf{Output Layer Distance Awareness}: Ensure that the output layer $g$ is distance aware, such that it produces an uncertainty metric reflecting the distance in the feature space $D_{\text{feature}}(f_\theta(I) - f_\theta(I'))$.
  \item \textbf{Feature Mapping Distance Preservation}: Ensure that the feature extractor $f_\theta$ is distance-preserving, meaning $D_F(f_\theta(I) - f_\theta(I'))$ corresponds meaningfully to the distance $D_{\text{input}}(I - I')$ in the data manifold.
\end{enumerate}
This is equivalent to requiring the \textbf{bi-Lipschitz} constraint on the feature extractor $f_\theta$~\cite{liu2020simple}:
\begin{equation*}
   C_L D_\text{input}(x_1,x_2)\leq D_{\text{feat}}(f_\theta(x_1),f_\theta(x_2))\leq C_U D_{\text{input}}(x_1,x_2)
\end{equation*}
for all inputs $x_1$ and $x_2$, where $D_{input}$ and $D_{feat}$ denote distance metrics for the input and feature space, and $C_L$ and $C_U$ the lower and upper Lipschitz constants. 

The lower bound of the bi-Lipschitz constraint ensures sensitivity in the feature distance, ensuring that meaningful changes in the input data are reflected in the feature space.
The upper bound ensures smoothness in the features, preventing them from becoming too sensitive to semantically meaningless input changes.
A combination of the two constraints ensures that the learned representations maintain a robust and meaningful correspondence with the semantic properties of the input data~\cite{liu2020simple}.

Residual connections in Transformer or ResNet blocks satisfy the Lipschitz constraint, if the residual blocks do not amplify the differences in their inputs, which can be ensured if the weights of the residual blocks are Spectrally Normalized (SN)~\cite{miyato2018spectral}.
We treat the SurroundOcc backbone as if it satisfies the requirement for a lower bound, since it has the residual connections, and the model is sensitive to the input data.
To ensure smoothness in the feature space, we introduce a detection head composed of multiple MLP blocks with residual connections (Fig.~\ref{fig:method_overview}~\colorsquare{m_blue}) that are Spectrally Normalized~\cite{miyato2018spectral} during training to ensure the smoothness constraint.
In this way, we do not significantly modify the architecture, but adapt the model to produce useful uncertainty values.

\PAR{Epistemic and Aleatoric Uncertainty.}
To estimate the uncertainty of the distant-aware features, we follow the DDU method~\cite{mukhoti2023ddu} and use a Gaussian Mixture Model for density estimates (Fig.~\ref{fig:method_overview}~\colorsquare{m_red}).
The feature vectors are extracted from the penultimate linear layer of the prediction head, where each data point corresponds to a voxel in the 3D space.
To ensure a robust and representative feature space, we collect these feature vectors from the entire training dataset and fit a Gaussian Mixture Model (GMM) $q(y, z)$ with a single Gaussian component per class. %
This procedure requires neither OoD data nor additional training, and only necessitates a single forward pass on the training set.
At test time, epistemic uncertainty is estimated by evaluating the log-likelihood of a data sample under the probability density function defined by the fitted GMM.
A high feature-space density indicates low epistemic uncertainty, making the aleatoric estimate from the softmax entropy reliable.
Conversely, a low feature density indicates high epistemic uncertainty, rendering the softmax prediction unreliable~\cite{mukhoti2023ddu}.

\begin{table*}[ht]
\centering
\caption{
\textbf{Quantitative evaluation of our proposed method against the established baselines.} 
We report accuracy metrics (IoU and mIoU), epistemic uncertainty quality proxies (mAUROC and mFPR95) and model complexity indicators. mAUROC and mFPR95 are computed across all image corruptions and all severity levels.
Our method demonstrates superior performance in terms of epistemic uncertainty quantification while maintaining the properties of the unmodified base model.
}
\begin{tabularx}{\linewidth}{lXXXXXXXXccc}
\toprule
\multirow{2}{*}    & \multicolumn{2}{c}{\thead{nuScenes}} & \multicolumn{2}{c}{\thead{MultiCorrupt}} & \multicolumn{2}{c}{\thead{Scene Corruptions}} & \multicolumn{2}{c}{\thead{Region Corruptions}} \\
\cmidrule(lr){2-3} \cmidrule(lr){4-5}  \cmidrule(lr){6-7}  \cmidrule(lr){8-9}
Method             & IoU$\uparrow$ & mIoU$\uparrow$ & IoU$\uparrow$ & mIoU$\uparrow$ & mAUROC$\uparrow$ & mFPR95$\downarrow$ & mAUROC$\uparrow$ & mFPR95$\downarrow$ & Params$\downarrow$ & Time(s)$\downarrow$ & Memory$\downarrow$ \\
\midrule
$\text{MCD}_{n=5}$ & 0.331             & 0.209             & 0.242              & 0.137             & 63.03            & 87.44            & 51.21          & 94.45  & \textbf{180.07M} & 2.11            & \textbf{7.1GB}  \\ 
$\text{DE}_{n=3}$  & \underline{0.338} & \underline{0.218} & \underline{0.246 } & \underline{0.145} & 68.46            & 76.64            & 54.26          & \underline{94,16}  & 540.22M          & 1.26            & 21.3GB \\ 
$\text{DE}_{n=5}$  &\textbf{0.341}     & \textbf{0.221}    & \textbf{0.247}     & \textbf{0.148}    & \underline{69.58} & \underline{74.89}& \underline{55.26} & 94.23  & 900.37M          & 2.11            & 35.5GB \\
\midrule
Entropy            & 0.328        & 0.208                  & 0.242 & 0,141 & 54.12         & 95.06             & 50.01          & 94.32 & \textbf{180.51M}    & \textbf{0.42}   & \textbf{7.1GB} \\
Max. Softmax       & 0.328        & 0.208                  & 0.242 & 0,141 & 55.42         & 87.71             & 50.22          & 95.65 & \textbf{180.51M}    & \textbf{0.42}   & \textbf{7.1GB} \\
Ours               & 0.329        & 0.210                  & 0.242 & 0.141 & \textbf{80.17}& \textbf{56.39} & \textbf{57.81} & \textbf{93.60 }& \underline{180.51M} & \underline{0.45}& \underline{7.1GB} \\
\bottomrule
\end{tabularx}
\label{tab:main_results}
\end{table*}

\section{Experiments}
We evaluate the effectiveness of our approach in capturing epistemic uncertainty using the nuScenes and a corrupted version of the nuScenes dataset~\cite{beemelmanns2024multicorrupt}.
Due to the unavailability of voxel-level OoD data, we establish an OoD detection framework at scene and region level.
In this context, our approach consistently outperforms Monte Carlo Dropout (MCD) and Deep Ensembles (DE).

\subsection{Experimental Setup}
We use the nuScenes dataset~\cite{caesar2020nuscenes} for In-Distribution (ID) training and evaluation.
The dataset consists of a diverse set of urban scenarios across multiple cities, where each frame includes six multi-view camera images, LiDAR and radar data.
For 3D occupancy prediction we use $6$ cameras as input, and semantically segmented LiDAR point clouds for the ground truth voxel labels.
The occupancy labels have a shape $200\times200\times16$ with a $0.5\text{m}$ voxel size.
Each voxel is labeled as one of the $17$ classes, including the \texttt{Unoccupied} class.
Our evaluations are based on the validation split that contains $6019$ scenes.
For training and evaluation of our method, we follow the SurroundOCC setup.

For OoD detection evaluations, we use MultiCorrupt~\cite{beemelmanns2024multicorrupt}, a corrupted version of the nuScenes dataset.
It contains common scene-level corruptions, such as fog, motion blur, or missing cameras, each at three levels of severity.
However, evaluating region-level uncertainties is a challenging task that requires voxel-level annotations.
It requires either the removal of the objects from the training set and training on the reduced set~\cite{bogdoll2022anomalya}, or expensive data collection.
To circumvent the problem of label scarcity, we evaluate region-level uncertainties by introducing corruptions for the frontal camera without corrupting the rest of the scene.

Following DDU~\cite{mukhoti2023ddu}, our Uncertainty Quantification Module requires a trained GMM model to estimate the feature density.
To fit the GMM, we randomly sample a maximum of $200$M feature vectors per class from the training set after occupancy prediction.
We compare our approach with three baseline methods: SurroundOCC base model with MLP head, but without SN and GMM, Monte Carlo Dropout (MCD) and Deep Ensembles (DE).
For MCD we enable dropout with $p=0.1$ at test time and perform $n=5$ forward passes.
For Deep Ensembles, we evaluate configurations with both $n=3$ and $n=5$ members.
For both MCD and DE, we compute the final predictions by averaging the softmax output across the multiple predictions and compute Predictive Entropy (PE) as uncertainty measure.
We estimate the uncertainty for the base model using the maximum softmax probability and entropy.

\subsection{Out-of-Distribution Detection}
Following common practices in the community~\cite{kendall2017what, mukhoti2023ddu}, we assess the quality of the estimation of epistemic uncertainty by disentangling the ID and OoD data.
To evaluate ID and OoD separation, we compare corrupted with clean data samples and compute the commonly used AUROC and FPR95 metrics~\cite{yang2022openood}.
Due to the large problem size, we evaluate OoD detection on scene-level by comparing mean scene uncertainties.
We perform all OoD experiments on all three severity levels over five corruptions.
The experimental results are summarized in Table~\ref{tab:main_results}, where mAUROC and mFPR95 are the averages for all values of AUROC and FPR95 across all types of corruptions and severity levels, respectively.
Compared to the baselines, our method achieves higher scores in OoD detection at both scene and region level.

\subsection{Epistemic Uncertainty Estimates.}
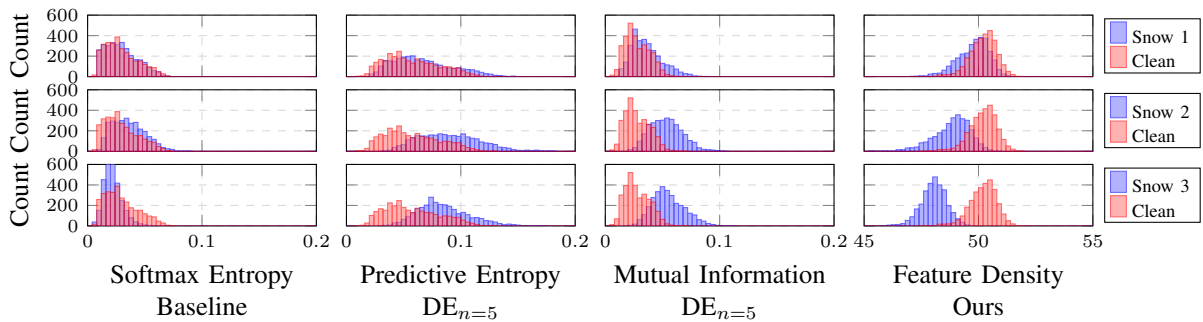
\begin{figure*}[ht]
\centering
\begin{tikzpicture}
\begin{groupplot}[
    group style={
        group size=4 by 3,
        horizontal sep=4mm,
        vertical sep=1.75mm,
        x descriptions at=edge bottom,
        y descriptions at=edge left
    },
    width=0.260\textwidth,
    height=2.4cm,
    ylabel={Count},
    ylabel near ticks,
    xmin=0, xmax=0.2, ymin=0, ymax=600,
    xtick={0.0, 0.1, 0.2},
    xlabel near ticks,
    xlabel style={align=center},
    xticklabel style={font=\scriptsize}, 
    yticklabel style={font=\scriptsize},
    grid style={dashed,gray!30},
    ymajorgrids=true,
    xmajorgrids=true,
    legend to name=donotshowthislegend,
    legend columns=1,
    legend style={fill=none, font=\scriptsize, inner xsep=2pt, inner ysep=1pt},
    title style={yshift=-0.23cm}
]

\nextgroupplot[]
\addplot+[
    label=l1,
    mark=no,
    fill=blue!50,
    opacity=0.6,
    hist={bins=50, data min=0, data max=0.2},
    ybar legend
] table [y index=0] {data/model2/softmax_entropy_snow1.csv};
\label{plot:brightness1}
\addplot+[
    mark=no,
    fill=red!50,
    opacity=0.6,
    hist={bins=50, data min=0, data max=0.2},
    ybar legend
] table [y index=0] {data/model2/softmax_entropy_clean.csv};
\label{plot:clean1}

\nextgroupplot[]
\addplot+[
    mark=no,
    fill=blue!50,
    opacity=0.6,
    hist={bins=50, data min=0, data max=0.2},
    ybar legend
] table [y index=0] {data/de5/pe_per_sample_snow1.csv};
\addplot+[
    mark=no,
    fill=red!50,
    opacity=0.6,
    hist={bins=50, data min=0, data max=0.2},
    ybar legend
] table [y index=0] {data/de5/pe_per_sample_clean.csv};

\nextgroupplot[]
\addplot+[
    mark=no,
    fill=blue!50,
    opacity=0.6,
    hist={bins=50, data min=0, data max=0.2},
    ybar legend
] table [y index=0] {data/de5/mi_per_sample_snow1.csv};
\addplot+[
    mark=no,
    fill=red!50,
    opacity=0.6,
    hist={bins=50, data min=0, data max=0.2},
    ybar legend
] table [y index=0] {data/de5/mi_per_sample_clean.csv};

\nextgroupplot[xmin=45, xmax=55, xtick={45.0, 50.0, 55.0},]
\addplot+[
    mark=no,
    fill=blue!50,
    opacity=0.6,
    hist={bins=50, data min=45, data max=55},
    ybar legend
] table [y index=0] {data/model2/logdensity_snow1.csv};
\addplot+[
    mark=no,
    fill=red!50,
    opacity=0.6,
    hist={bins=50, data min=45, data max=55},
    ybar legend
] table [y index=0] {data/model2/logdensity_clean.csv};

\nextgroupplot[]
\addplot+[
    mark=no,
    fill=blue!50,
    opacity=0.6,
    hist={bins=50, data min=0, data max=0.2},
    ybar legend
] table [y index=0] {data/model2/softmax_entropy_snow2.csv};
\label{plot:brightness2}
\addplot+[
    mark=no,
    fill=red!50,
    opacity=0.6,
    hist={bins=50, data min=0, data max=0.2},
    ybar legend
] table [y index=0] {data/model2/softmax_entropy_clean.csv};
\label{plot:clean2}

\nextgroupplot[]
\addplot+[
    mark=no,
    fill=blue!50,
    opacity=0.6,
    hist={bins=50, data min=0, data max=0.2},
    ybar legend
] table [y index=0] {data/de5/pe_per_sample_snow2.csv};
\addplot+[
    mark=no,
    fill=red!50,
    opacity=0.6,
    hist={bins=50, data min=0, data max=0.2},
    ybar legend
] table [y index=0] {data/de5/pe_per_sample_clean.csv};

\nextgroupplot[]
\addplot+[
    mark=no,
    fill=blue!50,
    opacity=0.6,
    hist={bins=50, data min=0, data max=0.2},
    ybar legend
] table [y index=0] {data/de5/mi_per_sample_snow2.csv};
\addplot+[
    mark=no,
    fill=red!50,
    opacity=0.6,
    hist={bins=50, data min=0, data max=0.2},
    ybar legend
] table [y index=0] {data/de5/mi_per_sample_clean.csv};

\nextgroupplot[xmin=45, xmax=55, xtick={45.0, 50.0, 55.0},]
\addplot+[
    mark=no,
    fill=blue!50,
    opacity=0.6,
    hist={bins=50, data min=45, data max=55},
    ybar legend
] table [y index=0] {data/model2/logdensity_snow2.csv};
\addplot+[
    mark=no,
    fill=red!50,
    opacity=0.6,
    hist={bins=50, data min=45, data max=55},
    ybar legend
] table [y index=0] {data/model2/logdensity_clean.csv};

\nextgroupplot[xlabel={Softmax Entropy \\ Baseline}]
\addplot+[
    mark=no,
    fill=blue!50,
    opacity=0.6,
    hist={bins=50, data min=0, data max=0.2},
    ybar legend
] table [y index=0] {data/model2/softmax_entropy_snow3.csv};
\label{plot:brightness3}
\addplot+[
    mark=no,
    fill=red!50,
    opacity=0.6,
    hist={bins=50, data min=0, data max=0.2},
    ybar legend
] table [y index=0] {data/model2/softmax_entropy_clean.csv};
\label{plot:clean3}

\nextgroupplot[xlabel={Predictive Entropy \\ $\text{DE}_{n=5}$}]
\addplot+[
    mark=no,
    fill=blue!50,
    opacity=0.6,
    hist={bins=50, data min=0, data max=0.2},
    ybar legend
] table [y index=0] {data/de5/pe_per_sample_snow3.csv};
\addplot+[
    mark=no,
    fill=red!50,
    opacity=0.6,
    hist={bins=50, data min=0, data max=0.2},
    ybar legend
] table [y index=0] {data/de5/pe_per_sample_clean.csv};

\nextgroupplot[xlabel={Mutual Information \\ $\text{DE}_{n=5}$}]
\addplot+[
    mark=no,
    fill=blue!50,
    opacity=0.6,
    hist={bins=50, data min=0, data max=0.2},
    ybar legend
] table [y index=0] {data/de5/mi_per_sample_snow3.csv};
\addplot+[
    mark=no,
    fill=red!50,
    opacity=0.6,
    hist={bins=50, data min=0, data max=0.2},
    ybar legend
] table [y index=0] {data/de5/mi_per_sample_clean.csv};

\nextgroupplot[xlabel={Feature Density \\ Ours}, xmin=45, xmax=55, xtick={45.0, 50.0, 55.0},]
\addplot+[
    mark=no,
    fill=blue!50,
    opacity=0.6,
    hist={bins=50, data min=45, data max=55},
    ybar legend
] table [y index=0] {data/model2/logdensity_snow3.csv};
\addplot+[
    mark=no,
    fill=red!50,
    opacity=0.6,
    hist={bins=50, data min=45, data max=55},
    ybar legend
] table [y index=0] {data/model2/logdensity_clean.csv};

\end{groupplot}
\matrix[draw, anchor=west, font=\scriptsize, inner xsep=2pt, inner ysep=1pt, /pgfplots/every crossref picture/.append style={yshift=-.75ex}] at ($(group c4r1.east)$) [right=0.15cm] {
  \ref{plot:brightness1}; & \node{Snow 1}; \\
  \ref{plot:clean1}; & \node{Clean}; \\
};
\matrix[draw, anchor=west, font=\scriptsize, inner xsep=2pt, inner ysep=1pt, /pgfplots/every crossref picture/.append style={yshift=-.75ex}] at ($(group c4r2.east)$) [right=0.15cm] {
  \ref{plot:brightness1}; & \node{Snow 2}; \\
  \ref{plot:clean1}; & \node{Clean}; \\
};
\matrix[draw, anchor=west, font=\scriptsize, inner xsep=2pt, inner ysep=1pt, /pgfplots/every crossref picture/.append style={yshift=-.75ex}] at ($(group c4r3.east)$) [right=0.15cm] {
  \ref{plot:brightness1}; & \node{Snow 3}; \\
  \ref{plot:clean1}; & \node{Clean}; \\
};

\end{tikzpicture}
\caption{\textbf{Feature density under data corruption.} The corruption type \emph{Snow} significantly impacts the feature density distribution of our proposed method leading to a near-complete separation between ID and OoD samples at severity level 3. In contrast, baseline approaches exhibit a less pronounced response to this shift in input distribution.}
\label{fig:brightness123_se_de5pe_de5mi_gmm}
\end{figure*}

We observe that the epistemic uncertainty, indicated by the log density, shows effective sensitivity across different types of corruption.
As an example, Fig.~\ref{fig:brightness123_se_de5pe_de5mi_gmm} illustrates the sensitivity of our approach compared to the baselines for corruption \emph{Snow}.

As the severity of the corruption increases, the feature density achieves better separation between the ID and OoD data.
This indicates that the feature density of our approach successfully quantifies epistemic uncertainty by assigning higher uncertainty values to unseen data.

\begin{table}[ht]
\caption{
\textbf{MLP Head Layers.}
We explore combinations of different number of MLP layers together with the use of skip connections in the UQ Module.
Our experiments support that skip connections are crucial to obtain sensitive features.
}
\centering
\begin{tabular}{cccccccc}
\toprule
Layers & Skip         & mAUROC$\uparrow$     & mFPR95$\downarrow$ & AP$\uparrow$  & Params$\downarrow$ \\
\midrule
3       &              & 0.770                & 0.618             &  0.760         & 180.336M \\
5       &              & 0.674                & 0.761             &  0.678         & 180.511M \\
3       & $\checkmark$ & 0.758                & 0.659             &  0.747         & 180.336M \\
5       & $\checkmark$ & \textbf{0.802}       & \textbf{0.564}    & \textbf{0.794} & 180.511M \\
\bottomrule
\end{tabular}
\label{table:mlp_head_ablation}
\end{table}

\subsection{UQ Module Design.}
Our uncertainty quantification module increases the inference time compared to the baseline model only slightly, while providing better estimates of epistemic uncertainty.
This efficiency makes it well-suited for real-time applications, in contrast to Deep Ensembles and Monte Carlo Dropout, which either require longer inference times or significantly larger memory footprint.

\begin{table}[ht]
\caption{
\textbf{Uncertainty Guided Temperature Scaling.} 
UGTS improves model calibration under corruptions by dynamically scaling logits based on uncertainty, unlike the standard fixed-temperature TS~\cite{ts}.
}
\centering
\begin{tabular}{m{0.85cm}K{0.60cm}K{0.60cm}cccc}
\toprule
\multirow{3}{*}  &              &      & \multicolumn{2}{c}{\thead{nuScenes}} & \multicolumn{2}{c}{\thead{MultiCorrupt}} \\
Method           & TS~\cite{ts} & UGTS & ECE$\downarrow$ & NLL$\downarrow$ & mECE$\downarrow$ & mNLL$\downarrow$  \\
\midrule
\multirow{3}{*}{$\text{DE}_{n=5}$} &      &          & 0.0388             & 0.284             & 0.0398             & 0.360            \\
                                   & $\checkmark$ &  & \underline{0.0165} & \underline{0.204} & 0.0193             & \underline{0.249} \\ 
                                   &  & $\checkmark$ & \textbf{0.0154}    & \textbf{0.203}    & \underline{0.0173} & \textbf{0.246} \\ 
                                   \cmidrule{2-7}
\multirow{3}{*}{Baseline}          &            &   & 0.0577             & 0.358             & 0.0646        & 0.461     \\
                                   & $\checkmark$ & & 0.0280             & 0.233             & 0.0350        & 0.288     \\
                                   & & $\checkmark$ & 0.0271             & 0.233             & 0.0343        & 0.287     \\ 
                                   \cmidrule{2-7}
\multirow{2}{*}{Ours}              & $\checkmark$ & & 0.0284             & 0.233             & 0.0352         & 0.288     \\
                                   & & $\checkmark$ & 0.0199             & 0.232             & \textbf{0.0171}& 0.282     \\
\bottomrule
\end{tabular}
\label{tab:temperature_scaling_results}
\end{table}

\PAR{Temperature Scaling}
In addition to the evaluation of uncertainty estimates, we report the model calibration on both corrupted and clean datasets.  
Inspired by recent work~\cite{yoon2024uncertainty}, we dynamically scale logits based on the sample's current uncertainty using temperature $t_{\text{new}}$,
\begin{equation}
    t_{\text{new}} = t_{\text{train}} \times \lambda (\bar{u}_{\text{sample}} - \bar{u}_{\text{train}})
\end{equation}
where $t_{\text{train}}$ is the optimal temperature and $\bar{u}_{\text{train}}$ the mean uncertainty on the train set for each method, respectively.
The difference between the training set mean and the current sample uncertainty $\bar{u}_{\text{sample}}$ is used to modulate the temperature $t_{\text{train}}$.
The parameter $\lambda$ accounts for the scale difference between the various uncertainty measures.
Hence, our Uncertainty Guided Temperature Scaling (UGTS) dynamically scales logits based on the current uncertainty.
We evaluate the Expected Calibration Error (ECE) and the Negative Log Likelihood (NLL) on the clean validation set and on MultiCorrupt, reporting the results in Table~\ref{tab:temperature_scaling_results}.
For a fair comparison, the parameter $\lambda$ is tuned for each UGTS model to achieve optimal results on the clean nuScenes validation split. For MultiCorrupt, we report mECE and mNLL, as mean values across all five corruptions and severity levels.
UGTS preserves model calibration, especially on corrupted data, without compromising accuracy on in-distribution data, at minimal additional cost.
\begin{table}[ht]
\caption{\textbf{OoD detection across output resolutions.} The highest resolution offers best OoD detection capabilities. Params denotes the number of additional parameters introduced by the GMM for 17 classes.}
\centering
\begin{tabular}{cccccc}
\toprule
Resolution & Dim.& mAUROC$\uparrow$ & mFPR95$\downarrow$ & Params$\downarrow$ \\
\midrule
20x20x2    & 256 & 58.00             & 90.81               & 1,118,464 \\
50x50x4    & 128 & 72.01             & 74.05               & 280,704 \\
100x100x8  & 64  & \underline{75.59} & \underline{69.93}   & \underline{69,649} \\
200x200x16 & 32  & \textbf{80.17}    & \textbf{56.39}      & \textbf{17,952} \\
\bottomrule
\end{tabular}
\label{table:output_scale_ablation}
\end{table}

\PAR{MLP Head Layers.}
A regularized feature space and skip connections within the detection head are essential to ensure smoothness and sensitivity, which is crucial to achieve distance awareness in the model.
We ablate the choice of the number of layers and skip connections, as shown in Table~\ref{table:mlp_head_ablation}.
For our experiments, we select the configuration that provides us with the best out-of-distribution performance, without increasing the number of parameters significantly.

\PAR{OoD performance at different scales.}
Occupancy predictors often impose supervision at multiple resolution scales during supervision. %
Our proposed evaluation setup with scene and region-level corruptions can be conceptually addressed at a coarser scale.
We explore four different output resolutions provided by the model and use them in the scene-level OoD detection task.
As illustrated in Table~\ref{table:output_scale_ablation}, finer resolutions consistently demonstrated superior separation between ID and OoD samples.
The results suggest that training the GMM with smaller and less ambiguous voxels contributes to a more accurate GMM model. %

\subsection{Qualitative Results}
\begin{figure}[ht]
  \centering
  \begin{subfigure}[b]{0.95\linewidth}
    \resizebox{\linewidth}{!}{%
    \begin{tikzpicture}[spy using outlines={yellow, magnification=2.0, height=20cm, width=15cm, connect spies}]
    \node {\includegraphics{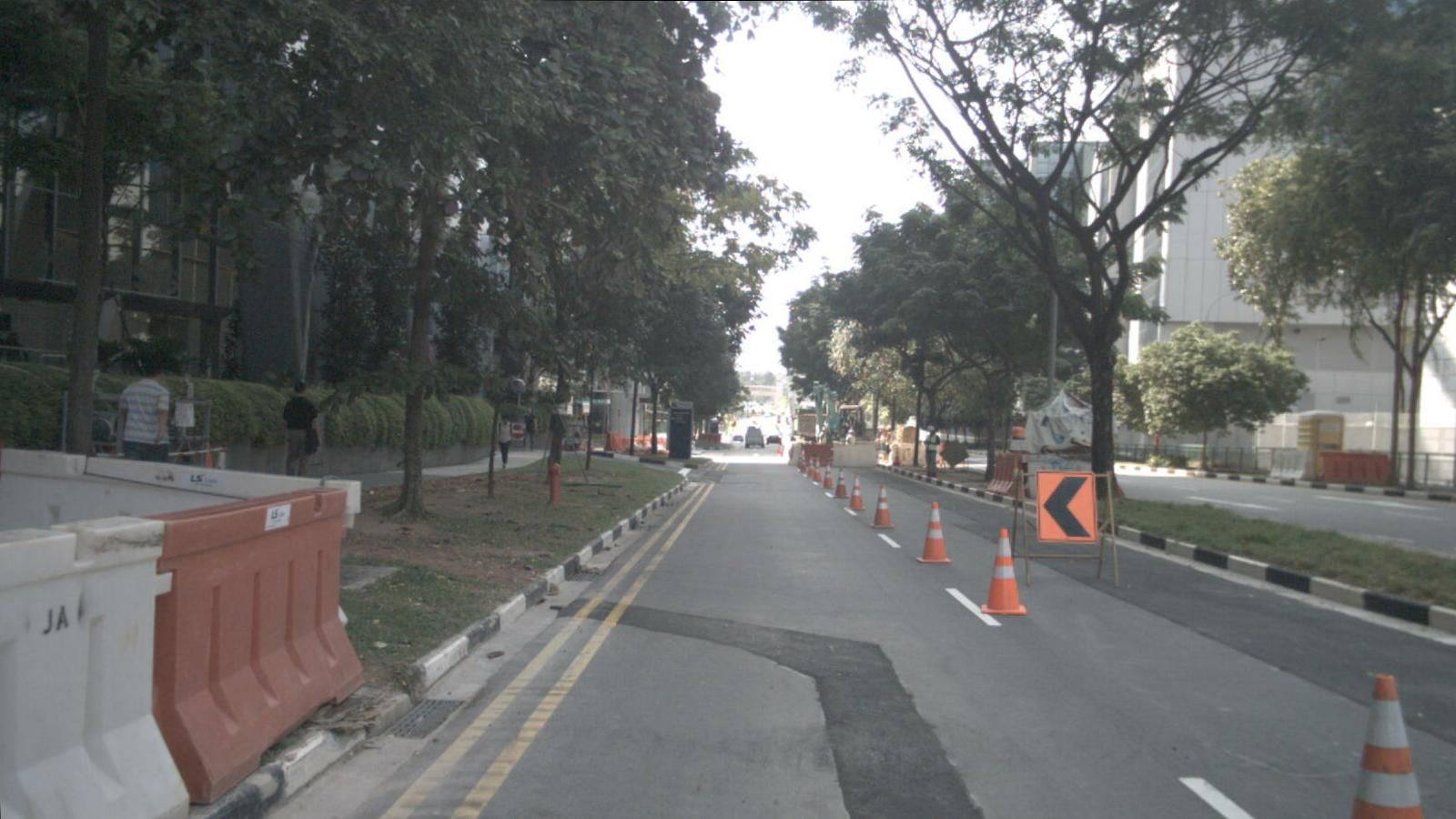}};
    \spy on (13.0,-4.0) in node [right] at (-26,5);
    \end{tikzpicture}%
  }%
  \caption{Front Center}\label{fig:sample952_front}%
  \end{subfigure}
  
  \begin{subfigure}[b]{0.49\linewidth}
    \resizebox{\linewidth}{!}{%
    \begin{tikzpicture}[spy using outlines={yellow, magnification=2.5, height=3cm, width=3cm, connect spies}]
    \node {\includegraphics[trim={4cm 2cm 0cm 2cm},clip]{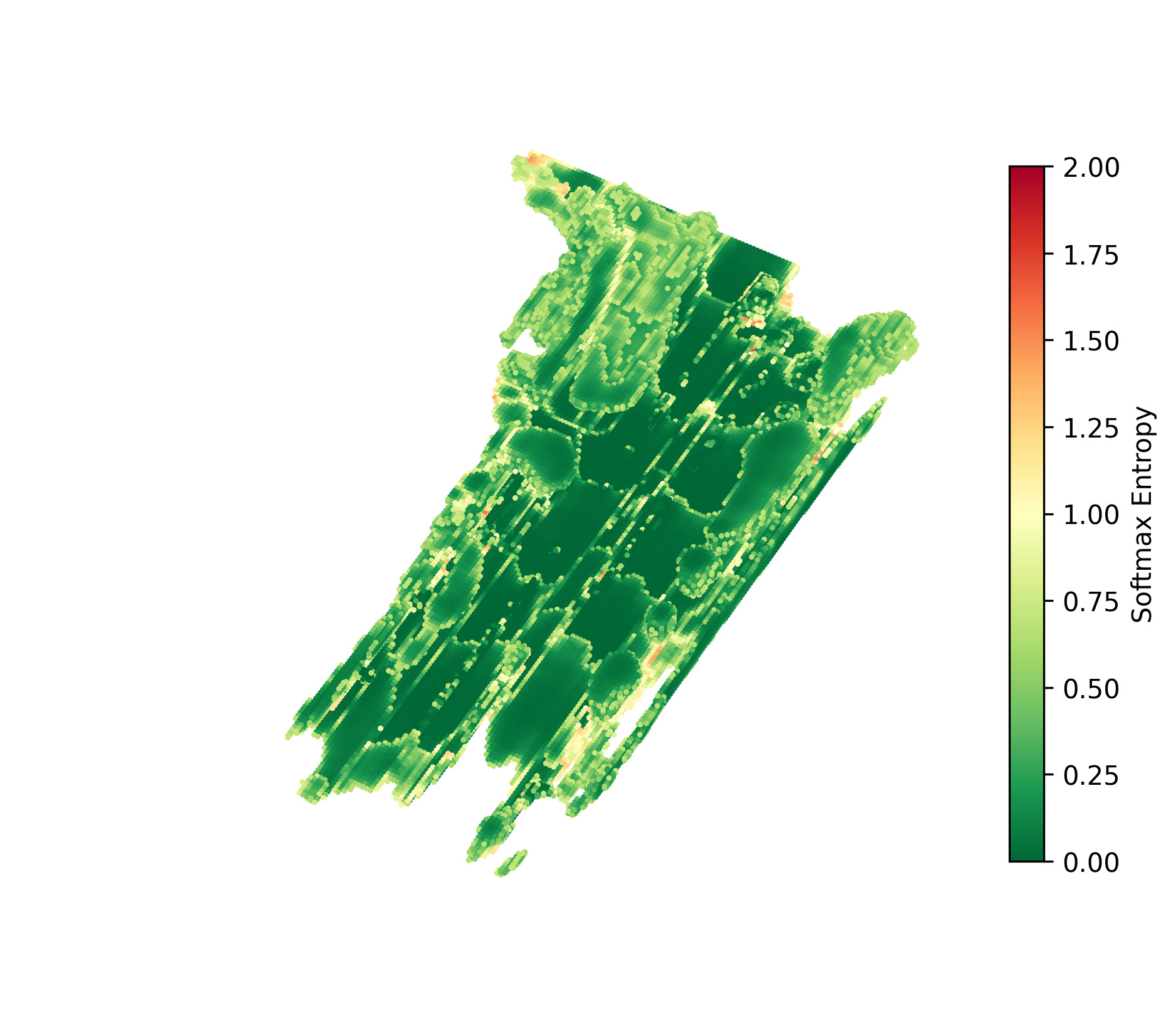}};
    \spy on (-3.0,-1.0) in node [right] at (-6,3.5);
    \end{tikzpicture}%
  }%
  \caption{Softmax Entropy}\label{fig:sample952_se}%
  \end{subfigure}%
  \begin{subfigure}[b]{0.49\linewidth}
    \resizebox{\linewidth}{!}{%
    \begin{tikzpicture}[spy using outlines={yellow, magnification=2.5, height=3cm, width=3cm, connect spies}]
    \node {\includegraphics[trim={4cm 2cm 0cm 2cm},clip]{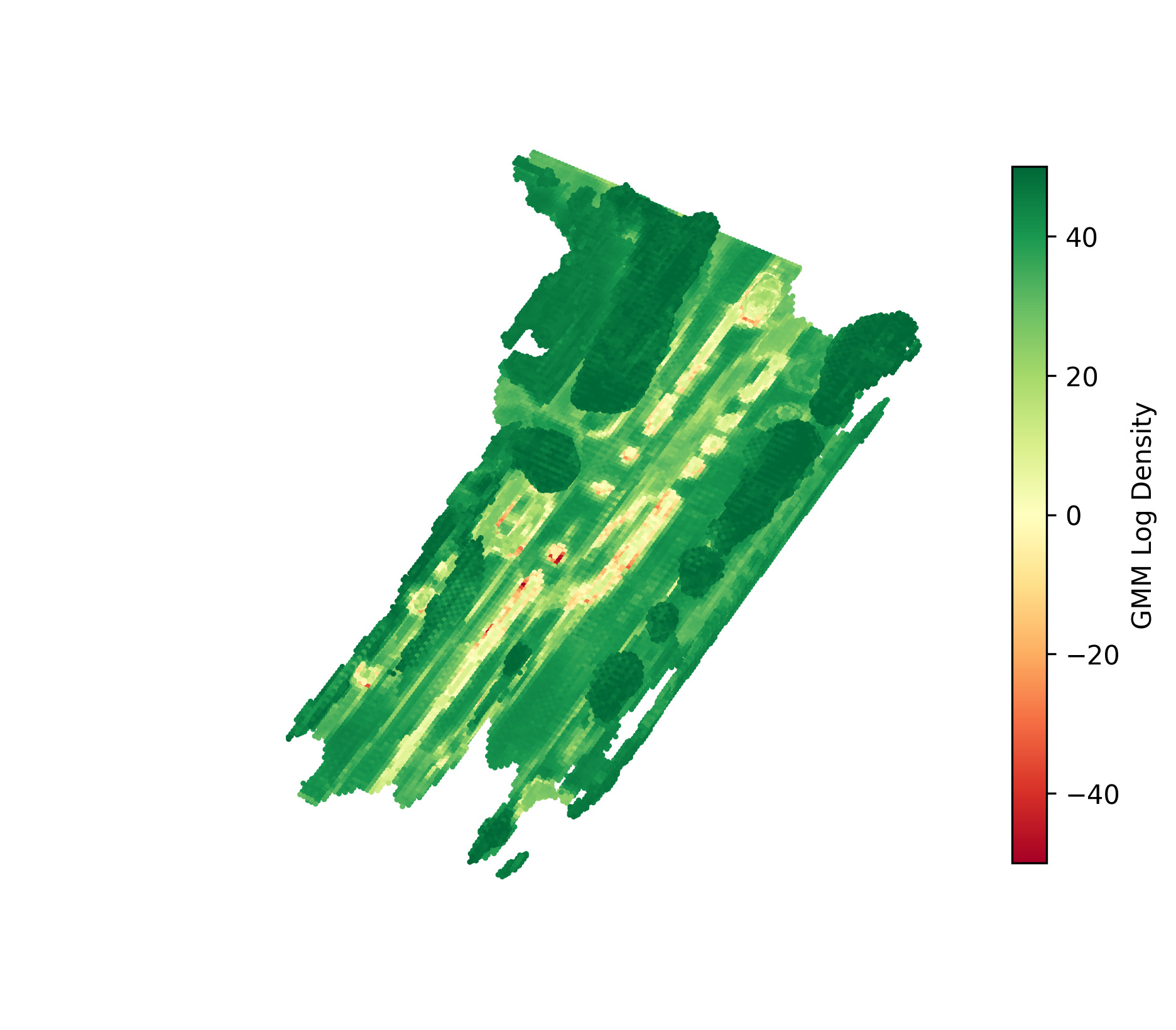}};
    \spy on (-3.0,-1.0) in node [right] at (-6,3.5);
    \end{tikzpicture}%
  }%
  \caption{Feature Density}\label{fig:sample952_feature_density}
  \end{subfigure}
  \begin{subfigure}[b]{0.49\linewidth}
    \resizebox{\linewidth}{!}{%
    \begin{tikzpicture}[spy using outlines={yellow, magnification=2.5, height=3cm, width=3cm, connect spies}]
    \node {\includegraphics[trim={4cm 2cm 0cm 2cm},clip]{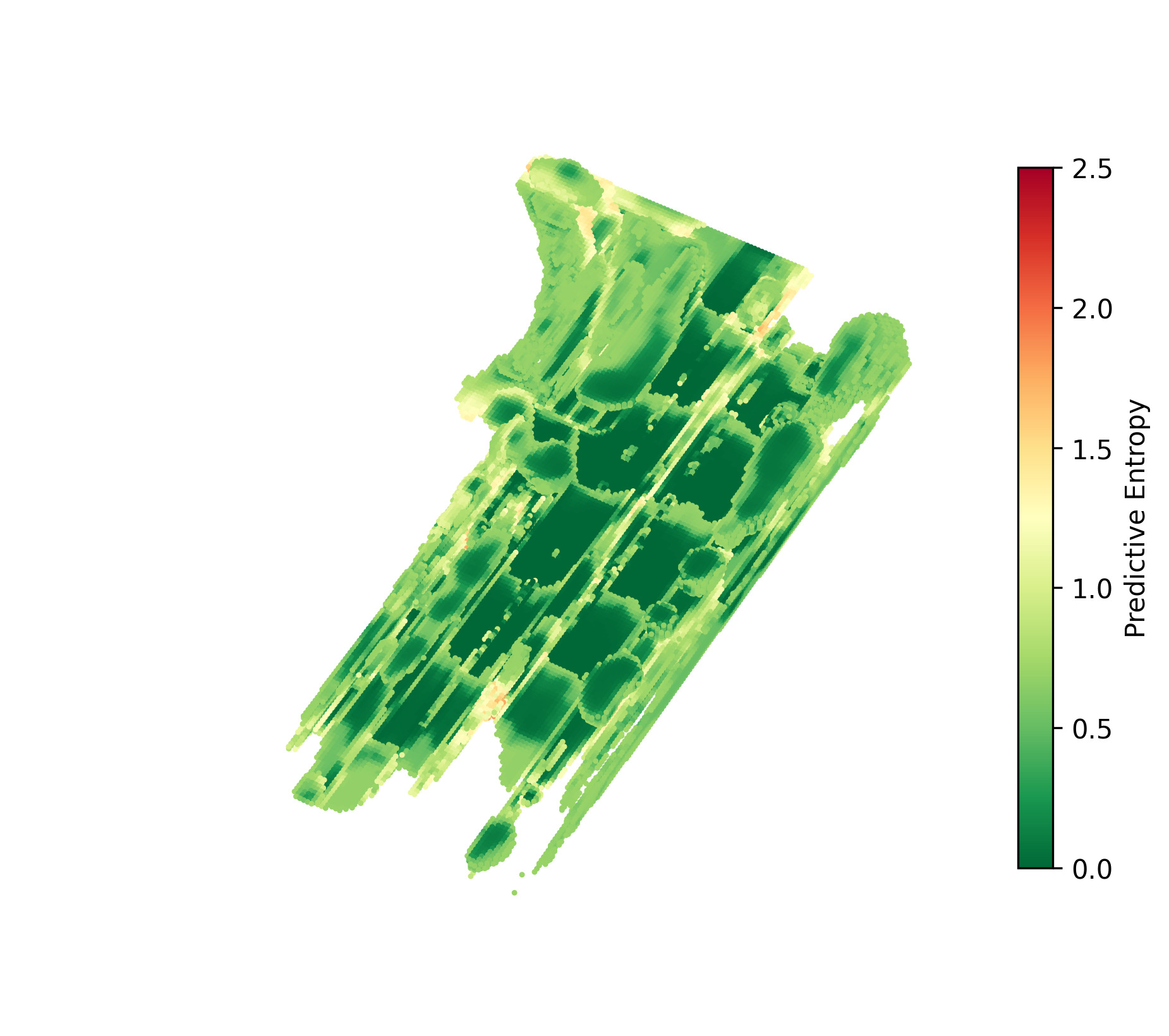}};
    \spy on (-3.0,-1.0) in node [right] at (-6,3.5);
    \end{tikzpicture}%
  }%
  \caption{$\text{DE}_{n=5}$ PE}\label{fig:sample952_de5pe}
  \end{subfigure}%
  \begin{subfigure}[b]{0.49\linewidth}
    \resizebox{\linewidth}{!}{%
    \begin{tikzpicture}[spy using outlines={yellow, magnification=2.5, height=3cm, width=3cm, connect spies}]
    \node {\includegraphics[trim={4cm 2cm 0cm 2cm},clip]{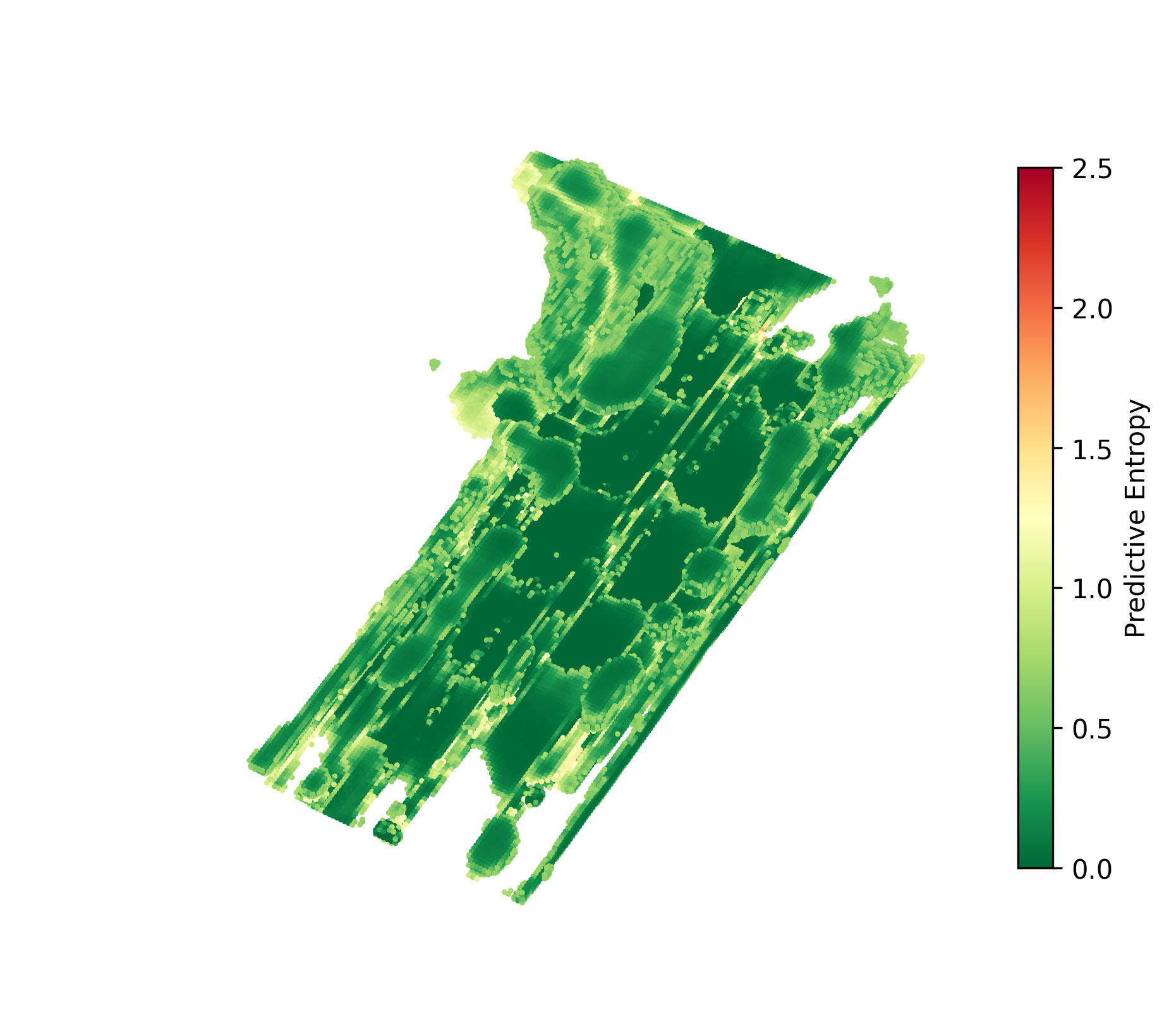}};
    \spy on (-3.0,-1.0) in node [right] at (-6,3.5);
    \end{tikzpicture}%
  }%
  \caption{$\text{MCD}_{n=5}$ PE}\label{fig:sample952_mcd5_pe}
  \end{subfigure}%
  \caption{\textbf{Construction site scenario.} The front camera (\subref{fig:sample952_front}) captures a challenging scene with numerous traffic cones, rare traffic sign objects, nearby a construction site. While baseline models (\subref{fig:sample952_se}), (\subref{fig:sample952_de5pe}) and (\subref{fig:sample952_mcd5_pe}) fail to present the uncertainty in this scenario, the feature density (\subref{fig:sample952_feature_density}) reveals high uncertainty near and around these objects, including the construction site.}
  \label{fig:sample952}
\end{figure}

We explore individual cases instead of region level corruptions to confirm the transfer of the scene and region uncertainty to voxel level.
The feature space density along with other baselines is shown in Fig.~\ref{fig:sample952}.
The feature density reveals high uncertainty for less frequently seen objects such as construction sites with rare road signs.
This is consistent with the definition of epistemic uncertainty, which is elevated for previously unseen or rare objects.

\section{Conclusion}
This work presents an efficient approach to 3D occupancy prediction that incorporates Uncertainty Quantification to enhance the reliability and safety of autonomous driving systems.
By adapting the DDU method for voxel-wise uncertainty estimation, we effectively address the limitations of existing methods that often neglect uncertainty quantification.
The proposed method effectively identifies corrupted data samples at both scene and region levels, without compromising prediction performance.
Moreover, the additional epistemic uncertainty estimate offers a means to dynamically calibrate the model's prediction confidence.
The proposed approach provides a lightweight modification with little computational overhead.

Future work should also focus on developing datasets with voxel-level OoD data to enable a more comprehensive evaluation of epistemic uncertainty and voxel-level OoD detection capabilities.
By addressing these areas, this research lays a strong foundation for advancing the field of 3D occupancy prediction and ensuring the safety and reliability of autonomous vehicles.

\bibliographystyle{IEEEtran}
\bibliography{root}

\clearpage

\end{document}